\documentclass{mathincs}
\usepackage{graphicx}
\usepackage{url}
\usepackage{xspace}
\usepackage{hyperref}
\usepackage{color}
\usepackage{array}
\usepackage{underscore}
\usepackage{tipa}
\usepackage{cite}
\usepackage{stmaryrd}
\usepackage{amsmath}
\usepackage{amssymb}
\usepackage{booktabs}
\usepackage{listings}
\usepackage{paralist}
\usepackage{subfig}
\usepackage{hyperref}
\usepackage{breakurl}
\usepackage{tikz}
\usepackage{textcomp}
\usepackage{alltt}
\usepackage{algorithm}
\usepackage{algpseudocode}

\usepackage{footnote}

\makeatletter

\def\holdocspecials{\do\ \do\$\do\&%
  \do\#\do\^\do\^^K\do\_\do\^^A\do\%}

\def\holtt{\trivlist \item[]\if@minipage\else\vskip\parskip\fi
\leftskip\@totalleftmargin\rightskip\z@
\parindent\z@\parfillskip\@flushglue\parskip\z@
\@tempswafalse \def\par{\if@tempswa\hbox{}\fi\@tempswatrue\@@par}
\obeylines \tt \let\do\@makeother \holdocspecials
 \frenchspacing\@vobeyspaces}

\makeatother

\newlength{\hsbw}
\newcommand\HOLSpacing{13pt}

\newenvironment{holnb}{\begin{flushleft}
 \begin{minipage}[b]{\hsbw}
 \vspace*{.06in}
 \begingroup\small\baselineskip\HOLSpacing\footnotesize
 \begin{holtt}}{\end{holtt}\endgroup
 \end{minipage}
 \end{flushleft}}

   \newcommand\hilbert{\varepsilon}
   
   \newcommand{\Thm}{\(\vdash\)}
   \newcommand{\Cond}{\(\rightarrow\)}
   \newcommand{\Eqv}{\(\equiv\)}
   \newcommand{\Iff}{\(\Longleftrightarrow\)\hspace{-1.5mm}}
   \newcommand{\Fa}{\(\forall\)}
   \newcommand{\Et}{\(\exists\)}
   \newcommand{\Eu}{\(\exists_{unique}\)}

   \newcommand{\Impl}{\(\Longrightarrow\)\hspace{-1.5mm}}
   \newcommand{\Func}{\(\to\)\hspace{-1.5mm}}

   \newcommand{\Lam}{\(\lambda\)}
   
   \newcommand{\Minus}{\(-\)}
   \newcommand{\Lminus}{\(-\)\hspace{-1.5mm}}
   \newcommand{\Prime}{\('\)}
   \newcommand{\Und}{\_}
   \newcommand{\Lt}{\(<\)}
   \newcommand{\Gt}{\(>\)}
   \newcommand{\Leq}{\(\leq\)}
   \newcommand{\Geq}{\(\geq\)}
   \newcommand{\Eq}{\(=\)}
   \newcommand{\Lrb}{\((\)}
   \newcommand{\Rrb}{\()\)}

   \newcommand{\Next}{\(\bigcirc\)}
   \newcommand{\Prev}{\(\ominus\)}
   \newcommand{\WPrev}{\(\widetilde{\bigcirc}\)}
   \newcommand{\Event}{\(\Diamond\)}
   \newcommand{\Once}{\(\underline{\Diamond}\)}  
\newcommand{\Hilbert}{\(\hilbert\)}

\newcommand{\Conj}{\(\wedge\)}
\newcommand{\Disj}{\(\vee\)}
\newcommand{\Neg}{\(\neg\)}
\newcommand{\Pnd}{\(\Diamond\)}

\newcommand{\Models}{\(\models\)}

\long\def\holthm#1{{\def\Turns{\Thm} \rechol#1\end\end\end}}

\long\def\rechol#1#2#3{\let\next=\rechol\def\postnext{#2#3}\ifx#1\end
\let\next=\relax\def\postnext{\relax}
\else\ifx#1!\Fa                                          
\else\ifx#1@\Hilbert                                     
\else\ifx#1\#\Pnd                                        
\else\ifx#1'\Prime                                       
\else\ifx#1~\Neg                                         
\else\ifx#1\~\Neg
\else\ifx#1_\Und                                         
\else\ifx#1(\ifx#2+\ifx#3)\Next\def\postnext{}\fi        
            \else\ifx#2-\Prev\def\postnext{}             
            \else\ifx#2~\ifx#3)\WPrev\def\postnext{}\fi            
             \else\Lrb\fi\fi\fi                          
\else\ifx#1)\Rrb%
\else\ifx#1\/\Disj                                       
\else\ifx#1\.\Lam                                        
\else\ifx#1>\ifx#2=\Geq\def\postnext{#3}\else\Gt\fi      
\else\ifx#1?\ifx#2!\Eu\def\postnext{#3}\else\Et\fi       
\else\ifx#1-\ifx#2>\Func\def\postnext{#3}               
            \else\ifx#2-\Lminus\def\postnext{#3}
            \else\Minus\fi\fi                               
\else\ifx#1|\ifx#2-\Turns\def\postnext{#3}               
            \else\ifx#2=\Models\def\postnext{#3}
                 \else\Bar\fi\fi
\else\ifx#1<\ifx#2=\ifx#3>\Iff\def\postnext{}       
                   \else\Leq\def\postnext{#3}\fi    
            \else\ifx#2+\Event\def\postnext{}       
            \else\ifx#2-\Once\def\postnext{}       
            \else\Lt\fi\fi\fi                       
\else\ifx#1=\ifx#2=\ifx#3>\Impl\def\postnext{}            
                   \else\Eqv\def\postnext{#3}\fi         
            \else\ifx#2>\Cond\def\postnext{#3}
                 \else\Eq\fi\fi
\else\ifx#1/\ifx#2\^^M\Conj\par\def\postnext{#3}         
            \else\ifx#2\ \Conj\ \def\postnext{#3}\else#1\fi\fi  
\else#1\fi\fi\fi\fi\fi\fi\fi\fi\fi\fi\fi\fi\fi\fi\fi\fi\fi\fi\fi
\expandafter\next\postnext}

\newcolumntype{L}[1]{>{\raggedright\let\newline\\\arraybackslash\hspace{0pt}}m{#1}}

\def\systemname#1{\textsf{#1}\xspace}

\newcommand{\HOLLight}{\systemname{HOL Light}}
\newcommand{\HOL}{\systemname{HOL}}

\newcommand{\Isabelle}{\systemname{Isabelle}}
\newcommand{\Sledgehammer}{\systemname{Sledgehammer}}
\newcommand{\MetiTarski}{\systemname{MetiTarski}}
\newcommand{\Mizar}{\systemname{Mizar}}
\newcommand{\Vampire}{\systemname{Vampire}}
\newcommand{\Epar}{\systemname{Epar}}
\newcommand{\Z}{\systemname{Z3}}
\newcommand{\E}{\systemname{E}}

\newcommand{\MizAR}{\systemname{Miz$\mathbb{AR}$}}

\newcommand{\Flyspeck}{\systemname{Flyspeck}}

\newcommand{\MaLARea}{\systemname{MaLARea}}
\newcommand{\MaLeCoP}{\systemname{MaLeCoP}}

\newcommand{\OCaml}{\systemname{OCaml}}

\newcommand{\HH}{\systemname{HOL(y)Hammer}}
\newcommand{\Gitolite}{\systemname{Gitolite}}
\newcommand{\Git}{\systemname{Git}}
\newcommand{\Gitweb}{\systemname{Gitweb}}
\newcommand{\BliStr}{\systemname{BliStr}}

\title{\HH: Online ATP Service for \HOLLight}
\author{Cezary Kaliszyk}

\address{University of Innsbruck, Austria}
\author{Josef Urban}

\address{Radboud University, Nijmegen}

\begin{document}
\maketitle

\begin{abstract}
  \HH is an online AI/ATP service for formal (computer-understandable)
  mathematics encoded in the \HOLLight system. The service allows its
  users to upload and automatically process an arbitrary formal development (project) based on \HOLLight,
  and to attack arbitrary conjectures that use the concepts defined in
  some of the uploaded projects.
  For that, the service uses several automated reasoning systems
  combined with several premise selection methods trained on all the
  project proofs.  The projects that are readily available on the server for such
  query answering include the recent versions of the Flyspeck,
  Multivariate Analysis and Complex Analysis libraries.
  The service runs on a 48-CPU
  server, currently employing in parallel for each task 7 AI/ATP
  combinations and 4 decision procedures that contribute to its
  overall performance. The system is also available for local
  installation by interested users, who can customize it for their own
  proof development. An Emacs interface allowing parallel asynchronous
  queries to the service is also provided.  The overall structure of
  the service is outlined, problems that arise and their solutions are discussed, and an
  initial account of using the system is given.
\end{abstract}

\section{Introduction and Motivation}
\label{Introduction}
\HOLLight~\cite{Harrison96} is one of the best-known interactive
theorem proving (ITP) systems. It has been used to prove a number of
well-known mathematical
theorems\footnote{\url{http://www.cs.ru.nl/~freek/100/}} and as a platform
for formalizing the proof of the Kepler conjecture targeted by the Flyspeck
project~\cite{Hales05}. The whole \Flyspeck development, together
with the required parts of the \HOLLight library consisted of about
14.000 theorems as of June 2012, growing to about 19.000 theorems as of August 2013. Motivated by the development of
large-theory automated theorem
proving~\cite{Urban11-ate,HoderV11,sledgehammer10,US+08} and its
growing use for ITPs like Isabelle~\cite{PaulsonS07} and
Mizar~\cite{abs-1109-0616,UrbanS10}, we have recently implemented
translations from \HOLLight to ATP (automated theorem proving)
formats, developed a number of premise-selection techniques for
\HOLLight, and experimented with the strongest and most orthogonal
combinations of the premise-selection methods and various ATPs. This
initial work, described in~\cite{holyhammer}, has shown that 39\% of the (June 2012) 14185
\Flyspeck theorems could be proved in a push-button mode (without any
high-level advice and user interaction) in 30 seconds of real time on
a fourteen-CPU workstation. More recent work on the AI/ATP methods have 
raised this performance to 47\%~\cite{EasyChair:74}.

The experiments that we did emulated the \Flyspeck development (when
the user always knows all the previous proofs\footnote{The Flyspeck processing order is used to define precisely what ``previous'' means. See~\cite{holyhammer} for details.} at a given point, and wants to
prove the next theorem), however they were all done in an offline mode
which is suitable for such experimentally-driven research. The ATP
problems were created in large batches using different
premise-selection techniques and different ATP encodings (untyped
first-order~\cite{SutcliffeSCG06}, polymorphic typed
first-order~\cite{tff1}, and typed higher-order~\cite{GelderS06}), and
then attempted with different ATPs (17 in total) and different numbers
of the most relevant premises. Analysis of the results interleaved
with further improvements of the methods and data have
gradually led to the current strongest combination of the AI/ATP
methods.

This strongest combination now gives to a \HOLLight/\Flyspeck user 
a 47\% chance (when using 14 CPUs, each for 30s) that he will not
have to search the library for suitable lemmas and figure out the
proof of the next toplevel theorem by himself. For smaller (proof-local)
lemmas such likelihood should be correspondingly higher. To really
provide this strong automated advice to the users, the functions that
have been implemented for the experiments need to be combined into a
suitable AI/ATP tool. Our eventual
goal (from which we are of course still very far) should be an easy-to-use service, which in its online form offers
to formal mathematics (done here in \HOLLight, over the
concepts defined formally in the libraries) what services 
like Wolfram Alpha offer for informal/symbolic
mathematics. Some expectations, linked to the recent success of the
IBM Watson system, are today even
higher\footnote{See for example Jonathan Borwein's article: \url{http://theconversation.edu.au/if-i-had-a-blank-cheque-id-turn-ibms-watson-into-a-maths-genius-1213}}. 
Indeed, we believe that developing stronger and stronger AI/ATP tools
similar to the one presented here
is a necessary prerequisite providing the crucial semantic understanding/reasoning layer for building larger Watson-like systems
for mathematics that will (eventually) understand (nearly-)natural language
and (perhaps reasonably semanticized versions/alternatives of) \LaTeX{}.
The more
user-friendly and smarter such AI/ATP systems become, the higher also the chance
that mathematicians (and exact scientists) will get some nontrivial 
benefits\footnote{Formal verification itself is of course a great benefit, but its cost has been so far too high to attract most mathematicians.} 
from encoding mathematics (and exact science) directly in a computer-understandable form.

This paper describes such an AI/ATP service based on the formal
mathematical corpora like \Flyspeck developed with \HOLLight.  The
service -- \HH\footnote{See~\cite{Vinge92} for an example of future
  where AIs turn into deities. } (HH) -- is now available as a public
online system\footnote{\url{http://colo12-c703.uibk.ac.at/hh/}} instantiated for several large \HOLLight libraries,
running on a 48-CPU server spawning for each query by default 7 different AI/ATP
combinations and four decision procedures. We first describe in
Section~\ref{Online} the static (i.e., not user-updatable) problem solving
functions developed in the first simplified
version of the service for the most interesting
example of \Flyspeck. This initial version of the service allowed the
users to experiment with ATP queries over the fixed June 2012 version of
\Flyspeck for which the AI/ATP components had been gradually developed
over several months in the offline experiments described in~\cite{holyhammer}.
Section~\ref{Multi} then discusses the issues and solutions related to
running the service for multiple libraries and their versions at once,
allowing the users also to submit a new library to the server or to
update an existing library and all its AI/ATP components.
Section~\ref{Interaction} shows examples of interaction with the service,
using web, Emacs, and command-line interfaces.
The service can be also installed locally, and
trained on user's private developments. This is described in
Section~\ref{Local}. 
Section~\ref{Conclusion} concludes and
discusses future work.

\section{Description of the Problem Solving Functions for \Flyspeck}
\label{Online}

\begin{figure}[htb!]
\begin{center}
\input{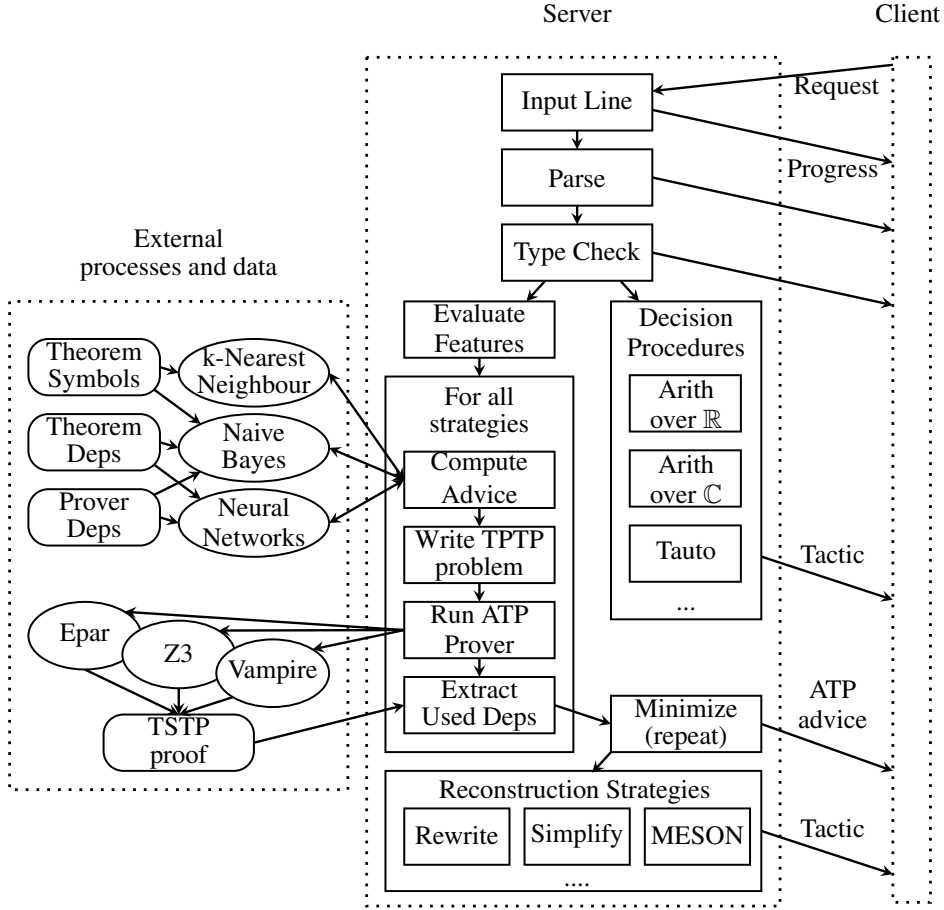}
\end{center}
\caption{Overview of the problem solving functions}\label{architecture}
\end{figure}

The overall problem solving architecture without the updating functions is shown in
Figure~\ref{architecture}. Since \Flyspeck is the largest and most interesting corpus on which this 
architecture was developed and tested, we use the \Flyspeck service as a running example in this whole section.
The service receives a query (a conjecture to
prove, possibly with local assumptions) generated by one of the
clients/frontends (Emacs, web interface, HOL session, etc.). If the query
produces a parsing (or type-checking) error, an exception is raised, and an error message
is sent as a reply. Otherwise the parsed query is processed in
parallel by the (time-limited) AI/ATP combinations and the native \HOLLight decision procedures 
(each managed by its forked
HOL Light process, and terminated/killed by the master process if not
finished within its global time limit). 
Each of the AI/ATP processes computes a specific feature
representation of the query, and sends such features to a specific
instance of a premise advisor trained (using the particular feature
representation) on previous proofs. Each of the advisors replies with
a specific number of premises, which are then translated to a
suitable ATP format, and written to a temporary file on which a
specific ATP is run. The successful ATP result is then (pseudo-)minimized,
and handed over to the combination of proof-reconstruction
procedures. These procedures again run in parallel, and if any of them
is successful, the result is sent as a particular tactic application
to the frontend. 
In case a native \HOLLight decision procedure finds a proof, the  
result (again a particular tactic application) can be immediately
 sent to the frontend. The following subsections explain this process in
more detail.

\subsection{Feature Extraction and Premise Selection}
\label{Features}
Given a (formal) mathematical conjecture, the selection of suitable
premises from a large formal library is an interesting AI problem, for
which a number of methods have been tried
recently~\cite{Urban11-ate,KuhlweinLTUH12,EasyChair:74}. The strongest methods use
machine learning on previous problems, combined in various ways with
heuristics like SInE~\cite{HoderV11}. To use the machine learning systems,
the previous problems have to be described as training examples in a
suitable format, typically as a set of (input) features characterizing
a given theorem, and a set of labels (output features) characterizing
the proof of the theorem. Devising good feature/label
characterizations for this task is again an interesting AI problem
(see, e.g.~\cite{UrbanS10}), however already the most obvious
characterizations like the conjecture symbols and the names of the
theorems used in the conjecture's proof are useful. This basic scheme
can be extended in various ways; see~\cite{holyhammer} for the
feature-extraction functions (basically adding various subterm and
type-based characteristics) and label-improving methods (e.g., using
minimized ATP proofs instead of the original \Flyspeck proofs whenever possible)
that we have so far used for
\HOLLight. For example, the currently most useful version of the characterization algorithm
would describe the \HOL theorem
\texttt{DISCRETE_IMP_CLOSED}:\footnote{\url{http://mws.cs.ru.nl/~mptp/hol-flyspeck/trunk/Multivariate/topology.html\#DISCRETE_IMP_CLOSED}}
\begin{holnb}\holthm{
!s:real^N->bool e.
        &0 < e /\ (!x y. x IN s /\ y IN s /\ norm(y - x) < e ==> y = x)
        ==> closed s
}\end{holnb}
by the following set of strings that encode its symbols and normalized types and terms:
\begin{holnb}
"real", "num", "fun", "cart", "bool", "vector_sub", "vector_norm",
"real_of_num", "real_lt", "closed", "_0", "NUMERAL", "IN", "=", "&0",
"&0 < Areal", "0", "Areal", "Areal^A", "Areal^A - Areal^A", 
"Areal^A IN Areal^A->bool", " Areal^A->bool", "_0", "closed Areal^A->bool",
"norm (Areal^A - Areal^A)", "norm (Areal^A - Areal^A) < Areal"
\end{holnb} 

On average, for each our feature-extraction method there are
in total about 30.000 possible conjecture-characterizing features
extracted from the theorems in the \Flyspeck development. The output
features (labels) are 
in the simplest setting just the names of the
\Flyspeck theorems\footnote{In practice, the
  \Flyspeck theorems are further preprocessed to provide better
  learning precision, for example by splitting conjunctions and
  detecting which of the conjuncts are relevant in which proof. Again,
 see~\cite{holyhammer} for the details. The number of labels 
used for the June 2012 \Flyspeck version with 14185 theorems is thus 16082.}  
extracted from the proofs with a modified (proof recording~\cite{newimport}) \HOLLight kernel.
These features and labels are (for
each extraction method) serially numbered in a stable way (using hashtables),
producing from all \Flyspeck proofs the training examples on which the
premise selectors are trained. The learning-based premise selection
methods currently used are those available in the
SNoW~\cite{Carlson1999} sparse learning toolkit (most prominently
sparse naive Bayes), together with a custom implementation~\cite{EasyChair:74} of the
distance-weighted $k$-nearest neighbor ($k$-NN) learner~\cite{DudaniS76}. Training a particular learning
method on all (14185) characterizations extracted from the \Flyspeck
proofs 
takes from 1 second for $k$-NN (a lazy
learner that essentially just loads all the 14185 proof
characterizations) and 6 seconds for naive Bayes using labels from minimized ATP
proofs, to 25 seconds for naive Bayes using the labels from the
original \Flyspeck proofs.\footnote{The original \Flyspeck proofs are often
using theorems that are in some sense redundant, resulting in longer
proof characterizations (and thus longer learning). This is typically a
consequence of using larger building blocks (e.g., decision
procedures, drawing in many dependencies) when constructing the ITP proofs.}
The trained premise selectors are then run as daemons (using their
server modes) that accept queries in the language of the numerical
features over which they have been trained, producing for each query their
ranking of all the labels, corresponding to the available \Flyspeck
theorems. 

Given a new conjecture, the first step of each of the forked \HOLLight
AI/ATP managing process is thus to compute the features of the
conjecture according to a particular feature extraction method,
compute (using the corresponding hashtable) the numerical
representation of the features, and send these numeric features as a
query to the corresponding premise-selection daemon. The daemon 
replies within a fraction of a second with its ranking, 
the exact speed depending on the learning method and the size of the feature/label sets.
This ranking is
translated back (using the corresponding table) to the
ranking of the \HOLLight theorems.
Each of the AI/ATP combinations then
uses its particular number (optimized so that the methods in the end complement
each other as much as possible) of the best-ranked theorems,
passing them together with the conjecture to the function that
translates such set of \HOLLight formulas to a suitable ATP format.

\subsection{Translation to ATP Formats and Running ATPs}

As mentioned in Section~\ref{Introduction}, several ATP formalisms are
used today by ATP and SMT systems. However the (jointly) most useful
proof-producing systems in our experiments turned out to be 
\E~\cite{Sch02-AICOMM} version 1.6 (run under the \Epar~\cite{blistr}
strategy scheduler), \Vampire~\cite{Vampire} 2.6, and  \Z~\cite{z3} 4.0. All
these systems accept the TPTP untyped first-order format (FOF).
Even when the input formalism (the \HOL logic~\cite{Pitts93} - polymorphic version of
Church's simple type theory) and the output formalism (TPTP FOF) are
fixed, there are in general many methods~\cite{BlanchetteBPS13} 
how to translate from the former
to the latter, each method providing different tradeoffs between soundness,
completeness, ATP efficiency, and the overall (i.e., including \HOL proof
reconstruction) efficiency. The particular method chosen by us
in~\cite{holyhammer} and used currently also for the service
is the polymorphic tagged
encoding~\cite{BlanchetteBPS13}. To summarize, the higher-order features (such as
lambda abstraction, application) of the \HOL formulas are first encoded
(in a potentially incomplete way) in first-order logic (still using
polymorphic types), and then type tags are added in a way that usually
guarantees type safety during the first-order proof search.

This translation method is in general not stable on the level of
single formulas, i.e., it is not possible to just keep in a global
hashtable the translated FOF version for each original \HOL formula,
as done for example for the \MizAR ATP service~\cite{abs-1109-0616}. This is because a
particular optimization (by Meng and Paulson~\cite{MengP08}) is used
for translating higher-order constants, creating for each such
constant $c$ a first-order function that has the minimum arity with
which $c$ is used in the particular set of \HOL formulas that is used
to create the ATP (FOF) problem. So once the particular AI/ATP
managing process advises its $N$ most-relevant \HOLLight theorems for
the conjecture, this
set of theorems and the conjecture are as a whole passed to the
translation function, which for each AI/ATP instance may produce
slightly different FOF encoding on the formula level. The encoding
function is
still reasonably fast, taking fractions of a second when using
hundreds of formulas, and still has the property that both the FOF
formula names and the FOF formulas (also those inferred during the ATP
proof search) can typically be decoded back into the original \HOL
names and formulas, allowing later \HOL proof reconstruction.

Each AI/ATP instance thus produces its specific temporary file (the
FOF ATP problem) and runs its
specific ATP system on it with its time limit. The time limit
is currently set globally to 30 seconds for each instance, however (as
usual in strategy scheduling setups) this could be made
instance-specific too, based on further analysis of the time
performance of the particular instances. \Vampire and \Epar already do such
scheduling internally: the current version of \Epar runs a fixed
schedule of 14 strategies, while \Vampire runs a problem-dependent
schedule of several to dozen of strategies. Assuming one strategy for
\Z and on average eight strategies for \Vampire, 
this now means that using 10-CPU parallelization results in about 100 different
proof-data/feature-extraction/learning/premise-slicing/ATP-strategy
instantiations tried by the online service within the 30 seconds of
the real time allowed for each query. Provided sufficient
complementarity of such instantiations and enough CPUs, this significantly raises
the overall power of the service.

\subsection{The AI/ATP Combinations Used}
An example of the 25 initially used combinations of the machine
learner, proof data, number of top premises used, the feature
extraction method, and the ATP system is shown in
Table~\ref{Combinations}. The proof data are either just the data from
the (minimized) ATP proofs (ATP0, ..., ATP3) created by a particular
(\MaLARea-style~\cite{US+08}, i.e., re-using the proofs found in
previous iteration for further learning) iteration of the
experimenting, possibly preferring either the \Vampire or \Epar proofs
(V_pref, E_pref), or a combination of such data from the ATP proofs
with the original \HOL proofs, obtained by slightly different versions
of the \HOL proof recording. Such combination typically uses the \HOL
proof only when the ATP proof is not available, see~\cite{holyhammer}
for details. The \texttt{standard} feature extraction method combines
the formula's symbols, standard-normalized subterms and normalized
types into its feature vector. The standard normalization here means
that each variable name is in each formula replaced by its normalized
\HOL type.  Types are normalized by renaming their polymorphic variables 
with de Bruijn indices. The \texttt{all-vars-same} and \texttt{all-vars-diff}
methods respectively just rename all formula variables into one common
variable, or keep them all different. This obviously influences the
concept of similarity used by the machine learners
(see~\cite{holyhammer} for more discussion). The 40-NN and 160-NN
learners are $k$-nearest-neighbors, run with $k=40$ and $k=160$.  The
particular combination of the AI/ATP is chosen by computing in a
greedy fashion the set of methods with the greatest coverage of the
solvable \Flyspeck problems. This changes often, whenever
some of the many components of this AI architecture get improved.  For
example, after the more recent strengthening of the premise-selection
and ATP components described in~\cite{EasyChair:74}, and the addition
of multiple developments and functions for their dynamic update
described in Section~\ref{Multi}, the number of AI/ATP combinations
run for a single query was reduced to 7.

\begin{table}[htb!]
\caption{The 25 AI/ATP combinations used by the initial \Flyspeck service}
\centering
\begin{tabular}{lcccc}
\toprule
Learner & Proofs   & Premises & Features  & ATP\\\midrule
Bayes & ATP2 & 0092 & standard & \Vampire \\
Bayes & ATP2 & 0128 & standard & \Epar \\
Bayes & ATP2 & 0154 & standard & \Epar \\
Bayes & ATP2 & 1024 & standard & \Epar \\
Bayes & HOL0+ATP0 & 0512 & all-vars-same & \Epar \\
Bayes & HOL0+ATP0 & 0128 & all-vars-diff & \Vampire \\
Bayes & ATP1 & 0032 & standard & \Z \\
Bayes & ATP1_V_pref & 0128 & all-vars-diff & \Epar \\
Bayes & ATP1_V_pref & 0128 & standard & \Z \\
Bayes & HOL0+ATP0 & 0032 & standard & \Z \\
Bayes & HOL0+ATP0 & 0154 & all-vars-same & \Epar \\
Bayes & HOL0+ATP0 & 0128 & standard & \Epar \\
Bayes & HOL0+ATP0 & 0128 & standard & \Vampire \\
Bayes & ATP1_E_pref & 0128 & standard & \Z \\
Bayes & ATP0_V_pref & 0154 & standard & \Vampire \\
40-NN & ATP1 & 0032 & standard & \Epar \\
160-NN & ATP1 & 0512 & standard & \Z \\
Bayes & HOL3+ATP3 & 0092 & standard & \Vampire \\
Bayes & HOL3+ATP3 & 0128 & standard & \Epar \\
Bayes & HOL3+ATP3 & 0154 & standard & \Epar \\
Bayes & HOL3+ATP3 & 1024 & standard & \Epar \\
Bayes & ATP3 & 0092 & standard & \Vampire \\
Bayes & ATP3 & 0128 & standard & \Epar \\
Bayes & ATP3 & 0154 & standard & \Epar \\
Bayes & ATP3 & 1024 & standard & \Epar \\\bottomrule
\end{tabular}
\label{Combinations}
\end{table}

\subsection{Use of Decision Procedures}

Some goals are hard for ATPs, but are easy for the existing decision
procedures already implemented in \HOLLight. To make the service more
powerful, we also try to directly use some of these \HOLLight decision
procedures on the given conjecture. A similar effect could be achieved
also by mapping some of the \HOLLight symbols (typically those
encoding arithmetics) to the symbols that are reserved and treated
specially by SMT solvers and ATP systems. This is now done for example
in \Isabelle/\Sledgehammer~\cite{sledgehammer10}, with the additional benefit of the
combined methods employed by SMTs and ATPs over various well-known
theories. Our approach is so far much simpler, which also means that we do
not have to ensure that the semantics of such special theories remains
the same (e.g., $1/0=0$ in \HOLLight). The \HOLLight decision
procedures might often not be powerful enough to prove whole theorems,
however for example the \texttt{REAL_ARITH}\footnote{\url{http://www.cl.cam.ac.uk/~jrh13/hol-light/HTML/REAL_ARITH.html}} tactic is called on 2678
unique (sub)goals in \Flyspeck, making such tools a useful addition to the service.

Each decision procedure is spawned in a separate instance of \HOLLight
using our parallel infrastructure, and if any returns within the timeout,
it is reported to the user. The decision procedures that we found most
useful for solving goals are:\footnote{The reader might wonder why
the above mentioned  \texttt{REAL_ARITH} is not among the tactics
used. The reason is that even though \texttt{REAL_ARITH} is used a lot
in \HOLLight formalizations, \texttt{INT_ARITH} is simply more
powerful. It solves 60\% more
\Flyspeck goals automatically without losing any of those
solved by \texttt{REAL_ARITH}. As with the AI/ATP instances, the
usage of decision procedures is optimized to jointly cover as many problems as
possible.}
\begin{itemize}
\item \texttt{TAUT}\footnote{\url{http://www.cl.cam.ac.uk/~jrh13/hol-light/HTML/TAUT.html}} --- Propositional tautologies.\\
  \verb|(A ==> B ==> C) ==> (A ==> B) ==> (A ==> C)|
\item \texttt{INT\_ARITH}\footnote{\url{http://www.cl.cam.ac.uk/~jrh13/hol-light/HTML/INT_ARITH.html}} --- Algebra and linear arithmetic over $\mathbb{Z}$ (including $\mathbb{R}$).\\
  \verb|&2 * &1 = &2 + &0|
\item \texttt{COMPLEX\_FIELD} --- Field tactic over $\mathbb{C}$ (including multivariate $\mathbb{R}$\footnote{\url{http://www.cl.cam.ac.uk/~jrh13/hol-light/HTML/REAL_FIELD.html}}).\\
  \verb|(Cx (&1) + Cx(&1)) = Cx(&2)|
\end{itemize}
Additionally the decision procedure infrastructure can be used to try
common tactics that could solve the goal. One that we found especially
useful is simplification with arithmetic (\texttt{SIMP\_TAC[ARITH]}), 
which solves a number of simple numerical goals that the service users
ask the server.

\subsection{Proof Minimization and Reconstruction}

When an ATP finds (and reports in its proof) a subset of the advised
premises that prove the goal, it is often the case that this set is
not minimal. By re-running the prover and other provers with only this set
of proof-relevant premises, it
is often possible to obtain a proof that uses less premises. A
common example are redundant equalities that may be used
by the ATP for early (but unnecessary) rewriting in the presence of
many premises, and avoided when the number of premises is
significantly lower (and different ordering is then used, or a completely
different strategy or ATP might find a very different proof).
This 
procedure is run recursively, until the number of premises needed for
the proof no longer decreases. We call this recursive procedure \textit{pseudo/cross-minimization}, 
since it is not exhaustive and uses multiple ATPs. Minimizing the number of premises
improves the chances of the \HOL proof reconstruction, and the speed
of (re-)processing large libraries that contain many such reconstruction
tactics.\footnote{Premise minimization has been for long
  time used to improve the quality and refactoring speed of the \Mizar
  articles. It is now also a standard part of \Sledgehammer.}

Given the minimized list of advised premises, we try to reconstruct the proof.
As mentioned in Section~\ref{Features}, the advice system may
internally use a number of theorem
names (now mostly produced by splitting conjunctions) not present in standard \HOLLight developments. It is possible to call the
reconstruction tactics with the names used internally in the advice system; however
this would create proof scripts that are not compatible with the original
developments. We could directly address the theorem sub-conjuncts
(using, e.g.,  ``\texttt{nth (CONJUNCTS thm) n}'')
however such proof scripts look quite unnatural (even if they are
indeed faster to process by \HOLLight). Instead, we now prefer
to use the whole original theorems (including all conjuncts) in the
reconstruction.

Three basic strategies are now tried to reconstruct the proof:
\texttt{REWRITE}\footnote{\url{http://www.cl.cam.ac.uk/~jrh13/hol-light/HTML/REWRITE_TAC.html}} (rewriting),
\texttt{SIMP}\footnote{\url{http://www.cl.cam.ac.uk/~jrh13/hol-light/HTML/SIMP_TAC.html}}
(conditional rewriting)
and \texttt{MESON}~\cite{Har96} (internal first-order ATP). 
These three strategies are started in parallel, each
with the list of \HOL theorems that correspond to the minimized list
of ATP premises as explained above.
The strongest of these tactics -- \texttt{MESON} -- can in one second reconstruct 79.3\%
of the minimized ATP proofs. While this is certainly useful, 
the performance of \texttt{MESON} reconstruction drops below 40\% as
soon as the ATP proof uses at least seven premises. Since the service
is getting stronger and stronger, the ratio of
\texttt{MESON}-reconstructable proofs is likely to get lower and lower.
That is why we have developed also a fine-grained reconstruction
method -- \texttt{HH_RECON}~\cite{proch}, which uses the quite
detailed TPTP proofs produced by \Vampire and \E. This method however
still needs an additional mechanism that maintains the TPTP proof as
part of the user development: either dedicated storage, or on-demand
ATP-recreation, or translation to a corresponding fine-grained
\HOLLight proof script. That is why \texttt{HH_RECON} is not yet
included by default in the service.

\subsection{Description of the Parallelization Infrastructure}

An important aspect of the online service is its parallelization
capability.  This is needed to efficiently process multiple requests
coming in from the clients, and to execute the large number of AI/ATP
instances in parallel within a short overall wall-clock time limit.
\HOLLight uses a number of imperative features of \OCaml, such as
static lists of constants and axioms, and a number of references
(mutable variables).  Also a number of procedures that are needed use
shared references internally. For example the \texttt{MESON} procedure
uses list references for variables. This makes \HOLLight not thread
safe.  Instead of spending lots of time on a thread-safe
re-implementation, the service just (in a pragmatic and simple way,
similar to the \Mizar parallelization~\cite{abs-1206-0141}) uses
separate processes (Unix fork), which is sufficient for our purposes. Given a list of \HOLLight tasks that
should be performed in parallel and a timeout, the managing process
spawns a child process for each of the tasks. It also creates a pipe
for communicating with each child process. Progress, failures or
completion information are sent over the pipe using \OCaml
marshalling.  This means that it is enough to have running just one managing
instance of \HOLLight loaded with \Flyspeck and with the advising
infrastructure. This process forks itself for each client query, and
the child then spawns as many AI/ATP, minimization, reconstruction, and decision
procedure instances as needed. 

\subsection{Use of Caching}

Even though the service can asynchronously process a number of
parallel requests, it is not immune to
overloading by a large number of requests coming in simultaneously. In
such cases, each response gets less CPU time and the requests are less
likely to succeed within the 30 seconds of wall-clock time. Such overloading is especially common for requests generated
automatically. For example the Wiki service that is being built for
Flyspeck~\cite{FlyspeckWiki} may ask many queries practically
simultaneously when an article in the wiki is
re-factored, but many of such queries will in practice overlap with previously
asked queries.  Caching is therefore employed by the service to efficiently serve such
repeated requests.

Since the parallel architecture uses different processes to serve
different requests, a file-system based cache is used (using file-level locking). For any
incoming request the first job done by the forked process handling the
request is to check whether an identical request has already
been served, and if so, the process just re-sends the previously computed answer. If
the request is not found in the cache, a new entry (file) for it is created,
and any information sent to the client (apart from the progress
information) is also written to the cache entry. This means that all
kinds of answers that have been sent to the client can be cached,
including information about terms that failed to parse or typecheck,
terms solved by ATP only, minimization results and replaying results,
including decision procedures.
The cache stored in the filesystem has the additional advantage of
persistence, and in case of updating the service the cache can
be easily invalidated by simply removing the cache entries.

\section{Multiple Projects, Versions, and Their Online Update}
\label{Multi}

The functions described in Section~\ref{Online} allowed the users to
experiment with ATP queries over the fixed June 2012 version of
\Flyspeck. If \Flyspeck already contained all of human mathematics in a
form that is universally agreed upon, such setting would be
sufficient. However, \Flyspeck is not the only library developed with
\HOLLight, and \Flyspeck itself has been updated considerably since
June 2012 with a number of new definitions, theorems and proofs. In
general, there is no final word on how formal mathematics should be
done, and even more stable formalization libraries may be updated,
refactored, and forked for new experiments.

\begin{figure}[thbp]
  \caption{The \HH web with a query over Multivariate Analysis}
\begin{center}
\includegraphics[width=13cm]{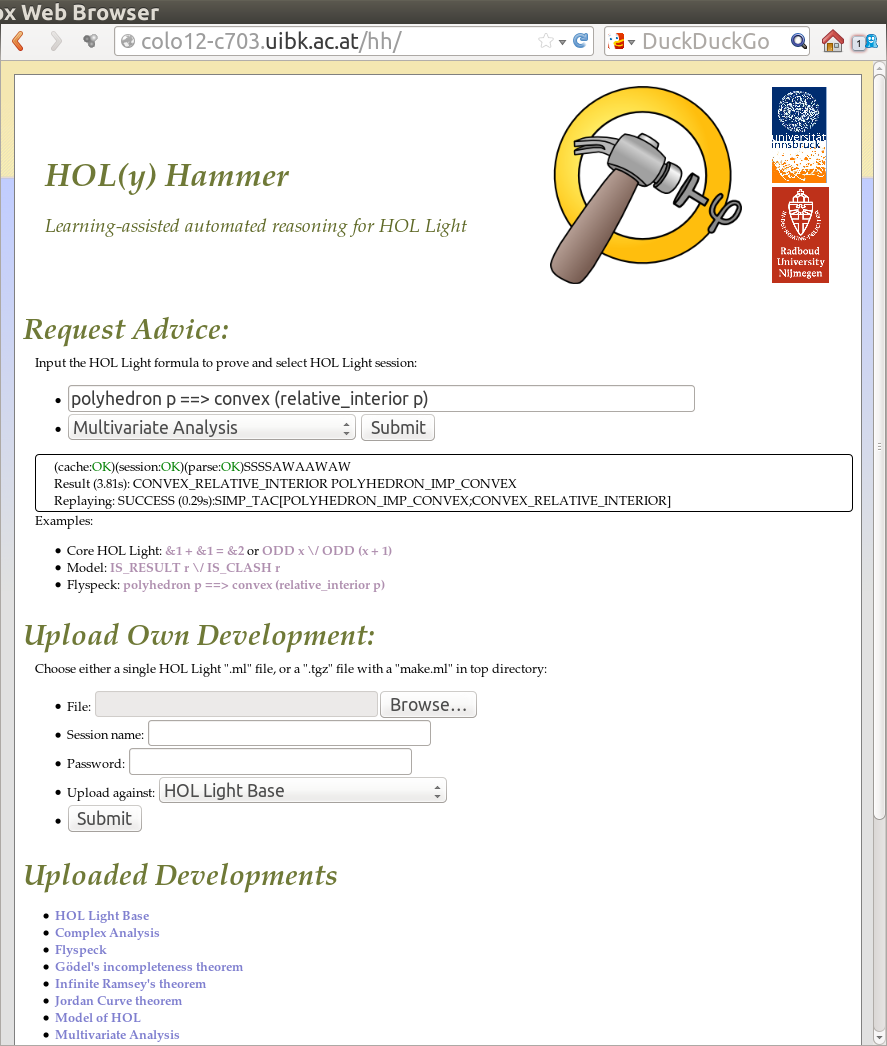}
\end{center}
\label{Web}
\end{figure}

To support this, the current version of \HH also allows
online addition of new projects and updating of existing
projects (see Figure~\ref{Web}). This leads to a number of issues that are discussed in this
section. A particularly interesting and important issue is the
transfer and re-use of the expensively obtained problem-solving
knowledge between the different projects and their versions.

Another major issue is the speed of loading large projects. \emph{Checkpointing} of \OCaml instances is used to save the load time,  
after \HOLLight was bootstrapped. Checkpointing software allows the
state of a process to be written to disk, and restore this state from
the stored image later. We use
\systemname{DMTCP}\footnote{\url{http://dmtcp.sourceforge.net}} as our
checkpointing software: it does not require kernel modifications, and
because of that it is one of the few checkpointing solutions that work
on recent Linux versions.

\subsection{Basic Server Infrastructure for Multiple Projects}
Instead of just one default project, the server allows multiple
projects identified by a unique project name such as ``Ramsey'',
``Flyspeck'' and ``Multivariate Analysis''. A new project can be
started by an authorized user performing a password-protected upload
of the project files via a HTTP POST request. In the same way, an
existing project can be updated.\footnote{\Git-based interface to the
  projects already exists and will probably also be used for updating
  the projects with the standard \texttt{git-push} command from users'
  computers. This still requires installation of the \Gitolite
  authentication layer on our server and implementing appropriate \Git
  hooks similar to those developed for the \Mizar wiki
  in~\cite{AlamaBMU11}.}.  The server data specific for each project are kept in
its separate directory, which includes
the user files, checkpointed images, features and proof dependencies used for learning premise selection, and the heuristically HTML-ized (hyperlinked) version of the user files. 
An overview of these project-specific data is given in Table~\ref{Data1}.

\begin{table}[htb!]
\caption{The data maintained for each \HH project.}
\begin{small}
\begin{tabular}{l L{12cm}}
Data & Description \\\toprule
  User files & User-submitted ML files . These data are additionally
  \Git-managed in this directory. \\
  Image1 & Checkpointed \HOLLight image preloaded with the user files and the HH functions. \\
  Image2 & An analogous image that uses proof recording to extract \HOL proof dependencies. \\
  Features & Several (currently six) feature characterizations (see Section~\ref{Features}) of the project's theorems. \\\hline
  \HOL deps & The theorem dependencies from the original \HOL proofs obtained by running the modified proof recording kernel on the user files. \\\hline
  ATP deps & The theorem dependencies obtained by running ATPs in various
  ways and minimizing such proofs. These data may be expensive to
  obtain, see~\ref{Reuse} for the current re-use mechanisms. \\\hline
  Cache & The request cache for the project. %
  \\
  Auxiliary& Auxiliary files useful for bookkeeping and debugging. \\\hline
  HTML &Heuristically HTML-ized version of the user files, together with
  index pages for the files and theorems. These files are available
  for browsing and they are also linked to the \Gitweb web interface,
  which presents the project and file history, allows comparison of different versions, regular
  expression search over the versions, etc. \\\bottomrule
\end{tabular}
\label{Data1}
\end{small}
\end{table}

Apart from the project-specific files, the service also keeps a spare %
checkpointed core \HOLLight image and additional files that typically
contain the reusable information from various projects. The core
\HOLLight image is used for faster creation of images for new
projects. A new project can also be cloned from an existing
project. In that case, instead of starting with the core \HOLLight
image, the new project starts with the cloned project's image, and loads
the new user files into them. This saves great amount of time when
updating large projects like \Flyspeck.
The server processing of a new or modified project is triggered by the
appropriate HTTP POST request. This starts the internal project
creator which performs on the server the actions described by Algorithm~\ref{Setup1}.
\begin{algorithm}[!htb]
\caption{Project creation stages}
\begin{small}
\begin{algorithmic}[1]
\State Set up the directory structure for new projects.
\State Open a copy of the checkpointed core \HOLLight image (or another project's cloned image) and load it with the user files and the \HH functions.
\State Export the typed and variable-normalized statements of named theorems together with their MD5 hashes.%
\State Export the various feature characterizations of the theorems.
\State Checkpoint the new image.
\State Re-process the user files with a proof-recording kernel that saves the (new) \HOL proof dependencies.
\State Checkpoint the proof-recording image.
\State Add further compatible proof dependencies from related projects.
\State Run ATPs on the problems corresponding to the \HOL dependencies, and minimize such proof data by running the ATPs further.
\State Run the heuristic HTML-izer and indexer, and push the user files to \Git.
\end{algorithmic}
\label{Setup1}
\end{small}
\end{algorithm}
The data sizes and processing times for seven existing projects are summarized in Table~\ref{Stats2} and Table~\ref{Stats3}.
\begin{table}[htb!]
\caption{The processing times for seven \HH projects in seconds.}
\centering
\begin{tabular}{lrrrrrrr}
& Core & Ramsey & Model & G\"odel & Complex & Multivariate & Flyspeck\\\toprule
Proof checking (min)& 3 & 6 & 193 &166 &267 &2716 & 21735\\
Proof recording (min)& 10 & 14 & 225 &215 &578 &3751 & 52002\\
Writing data & 26 & 27 & 33 &47 &53 &139 & 758\\
Writing ATP problems & 38.56 & 45.35 & 51.14 & 73.37 & 72.12 & 139.12 & 650.15 \\
Solving ATP problems & 1582.8 & 1622.4 & 1882.2 & 2173.8 & 2284.8 & 9286.2 & 12034.2 \\
HTML and Git & 4&2&2&3&2&19& 61\\
Image Restart &1.98&2.08&2.37&2.15&3.00&3.66&6.78\\
\end{tabular}
\label{Stats2}
\end{table}
\begin{table}[htb!]
\caption{The data sizes for seven \HH projects.}
\centering
\begin{tabular}{lrrrrrrr}
& Core & Ramsey & Model & G\"odel & Complex & Multivariate & Flyspeck\\\toprule
Normal image size (kB) &33892&40952	&37584	&38244	&55424	&77292	& 152460\\
Recording image size (kB) &50960&52692	&48148	&46000	&58368	&247848	& 365496\\
Unique theorems&2482&2544 &2951 &3408 &3582 &6798 &22336\\
Unique constants&234&234&337 &367 &333 &466 &1765\\
Avrg. \HOL proof deps.&12.13&12.27&11.09&14.44&17.96&12.26&21.86\\
ATP-proved theorems&1546&1578&1714&1830&2042&4126&8907\\
Usable ATP proofs &6094&6141&6419&6644&6885&11408&21733\\
Avrg. ATP proof deps.&6.86&6.86&6.77&6.67&6.94&6.36&6.52\\
Total distinct features&3735&3759&4693&5755&5964&11599&43858\\
Avrg. features/formula&24.81&24.61&26.05&35.61&39.05&38.15&67.61\\
\end{tabular}
{\small
  \begin{description}
\item[Usable ATP proofs:] \Vampire, \Epar and \Z are used, and we keep all the different minimal proofs. This means that the total number of ATP proofs can be higher than the number of theorems.
  \end{description}}
\label{Stats3}
\end{table}

\subsection{Safety} 
Since \HOLLight is implemented in the \OCaml
toplevel, allowing users to upload their own development is equivalent
to letting them run arbitrary programs on our server.\footnote{And
  indeed, the basic infrastructure could be also used as a platform
  for interacting with any \OCaml project.} This is inherently
insecure. A brief analysis of the related security issues and their
possible countermeasures has been done in the context of the
WorkingWiki~\cite{Worden:MathWiki11} collaborative
platform.\footnote{\url{http://lalashan.mcmaster.ca/theobio/projects/index.php/WorkingWiki/Security}}
The easiest practical solution is to allow uploads only by authorized
users, i.e., users who are sufficiently trusted to be given shell
access. Asking queries to existing projects can still be done
by anybody; the query is then just a string processed by a time-limited function
that always exits.

\subsection{Re-use of Knowledge from Related Projects}
\label{Reuse}
It has been shown in~\cite{holyhammer} that learning premise selection
from minimized ATP proofs is better than learning from the \HOL
proofs, and also that the two approaches can be productively combined,
resulting in further improvement of the overall ATP performance.
However, obtaining the data from ATP runs is expensive. For example,
just running \Vampire, \Epar and \Z on all \Flyspeck problems for 30
seconds takes (assuming 70\% unsolved problems for each ATP) about 500
CPU hours. Even with 50-fold parallelization, this takes 10 hours of
wall-clock time. And this is just the initial ATP
pass. In~\cite{holyhammer} we also show that further \MaLARea-style
learning from such ATP data and re-running of the ATPs with the
premises proposed by the learning grows the set of ATP solutions by
about 20\%. Obviously, such additional passes cost a lot of further
CPU time. 
One option is to sacrifice the ATP data for speed, and only learn from
the \HOL data, sacrificing the final ATP performance on the
queries. However, there is a relatively efficient way how to re-use
a lot of the expensive data that were already computed.  

Suppose that the user only updates an existing large project by adding
a new file. Then it is quite sufficient to (relatively quickly) obtain
the minimized ATP proofs of the (ATP-provable) theorems in the file
that was added. Such ATP proofs are then added to the existing training data
used for the premise selectors. In general, the project can however be
modified and updated in a more complicated way, for example by
adding/changing some files ``in the middle'', modifying symbol
definitions, theorems, etc. Or it can be a completely new project, that only
shares some parts with other projects, restructuring some terminology,
theorem names, and proofs. 
The method that we use to handle such cases efficiently is
\textit{recursive content-based encoding} of the theorem and symbol
names~\cite{UrbanMathWiki11}. This is the first practical
deployment and evaluation of this method, which in \HH is done as follows:

\begin{enumerate}
\item The name of every defined symbol is replaced by the content hash
  (we use \systemname{MD5}) of its variable-normalized definition
  containing the full types of the variables. This definition already
  uses content hashes instead of the previously defined symbols. This
  means that symbol names are no longer relevant in the whole project,
  neither white space and variable names.
\item The name of each theorem is also replaced by the content hash of
  its (analogously normalized) statement.
\item The proof-dependency data extracted in the content encoding
  from all projects are copied to a special ``common'' directory.
\item Whenever a project $P$ is started or modified, we find the
  intersection of the content-encoded names of the project's theorems
  with such names that already exist in other projects/versions.
\item For each of such ``already known'' theorems $T$ in $P$, we
  re-use all its ``already known'' proofs $D$ that are
  \textit{compatible} with $P$'s proof graph. This means, that the
  names of the proof dependencies of $T$ in $D$ must also exist in $P$
  (i.e., these theorems have been also proved in $P$, modulo the
  content-naming), and that these theorems precede $T$ in $P$ in its
  chronological order (otherwise we might get cyclic training data for
  $P$).
\end{enumerate}
There are two possible dangers with this approach: collisions in
\systemname{MD5} and dealing with types in the \HOL logic. The first
issue is theoretical: the chance of unintended \systemname{MD5}
collisions is very low, and if necessary, we can switch to stronger
hashes such as \systemname{SHA-256}. The second issue is more real:
there is a choice of using content-encoding also for the \HOL types,
or just using their original names. If original names are used, two
differently defined types can get the same name in two different
projects, making the theorems about such types incompatible. If
content encoding is used, all types with the same definition will get
the same content name. However, the \HOL logic rejects such semantic
equality of the two types already in its parsing layer: two
differently named types are always completely different in the \HOL
logic.\footnote{The second author could not resist pointing out that
  this issue disappears in set theory with soft types.} We currently
use the first method (keeping the original type names), however the
second method might be slightly more correct. In both cases, it
probably would not be hard to add guards against the possible
conflicts. In all cases, these issues only influence the performance
of the premise-selection algorithms. The theorem proving (and proof
reconstruction) is always done with the original symbols.

\subsection{Analysis of the Knowledge Re-use for \Flyspeck Versions}

It is interesting to know how much knowledge re-use can be obtained
with the content-encoding method. We analyze this in Table~\ref{ReuseT}
on the theorems (or rather unique conjuncts) coming from three
different \Flyspeck SVN versions: 2887, 3006, and 3400. Note that
the later versions have not been subjected to several learning/ATP
passes. Such passes raised the number of ATP-proved theorems in the
earlier versions by about 20\%. The table shows that the number of
reusable theorems and proofs from the previous version is typically
very high. This also means that more expensive AI/ATP computations (e.g., use of higher
time limits, \MaLARea-style looping, and even \BliStr-style strategy evolution~\cite{blistr})  could
be in the future added to the tasks done on the server in its idle
time, because the results of such computations will typically improve
the success rates of all the future versions of such large projects.

\begin{table}[htb!]
\caption{The re-use of theorems and ATP proofs between four \Flyspeck SVN versions}
\centering
\begin{tabular}{lccccc}
\toprule
Version & Unique thms
                  &In previous (\%) & ATP-proved (\%) & ATP proofs   & Reusable proofs (\%)\\\midrule
2887 &13647 & N/A         &7176 (53\%)& 20028 & N/A \\
3006 &13814 & 13480 (98\%)&7235 (52\%)& 20081 & 19977 (99\%)\\
3400 &18856 & 12866 (93\%)&8914 (47\%)& 21780 & 21320 (97\%)\\\bottomrule
\end{tabular}
{\small
  \begin{description}
  \item[In previous:] Theorems (conjuncts) that exist already in the previous version, and the per from.
  \item[ATP proof:] \Vampire, \Epar and \Z are used, and we keep all the different minimal proofs. This means that the total number of ATP proofs can be higher than the number of theorems.
  \item[Re-usable ATP proofs:] The proofs from the previous version that are valid also in the current version.
  \end{description}}
\label{ReuseT}
\end{table}

A by-product of the content encoding is also information about symbols
that are defined multiple times under different names. For the latest
version of \Flyspeck there are 39 of them, shown in Table~\ref{Symbols}

\begin{table}[htb!]
\begin{small}
\caption{39 symbols with the same content-based definition in \Flyspeck SVN 3400}
\centering
\begin{tabular}{l|c}
face_path / face_contour &   reflect_along / reflection	 \\
zero6 / dummy6		 &   UNIV / predT		\\	 
CROSS / *_c		 &   node3_y / rotate3		 \\
EMPTY / pred0		 &   APPEND / cat		\\	 
func / FUN		 &   set_components / set_part_components	  \\
ONE_ONE / injective	 &   triple_of_real3 / vector_to_pair \\
supp / SUPP		 &   is_no_double_joins / is_no_double_joints  \\
dirac_delta / delta_func &   unknown / NONLIN		 \\
o / compose		 &   node2_y / rotate2		 \\
I / LET_END / mark_term  & \\        			    
\end{tabular}
\label{Symbols}
\end{small}
\end{table}

\section{Modes of Interaction with the Service}
The standard web interface (Figure~\ref{Web}) displays the available
projects, links to their documentation, allows queries to the
projects, and provides an HTML form for uploading and modifying
projects.  Requests are processed using asynchronous DOM modification
(AJAX): a JavaScript script makes the requests in the background and
updates a part of the page that displays the response.  Each request
is first sent to the external PHP request processor, which
communicates with the \HH server. A prototype of a web editor
interacting both with \HOLLight and with the online advisor is
described in~\cite{FlyspeckWiki}.

\label{Interaction}
\begin{figure}[thbp]
  \caption{Parallel asynchronous calls of the online advisor from Emacs.}
\begin{center}
\includegraphics[width=10cm]{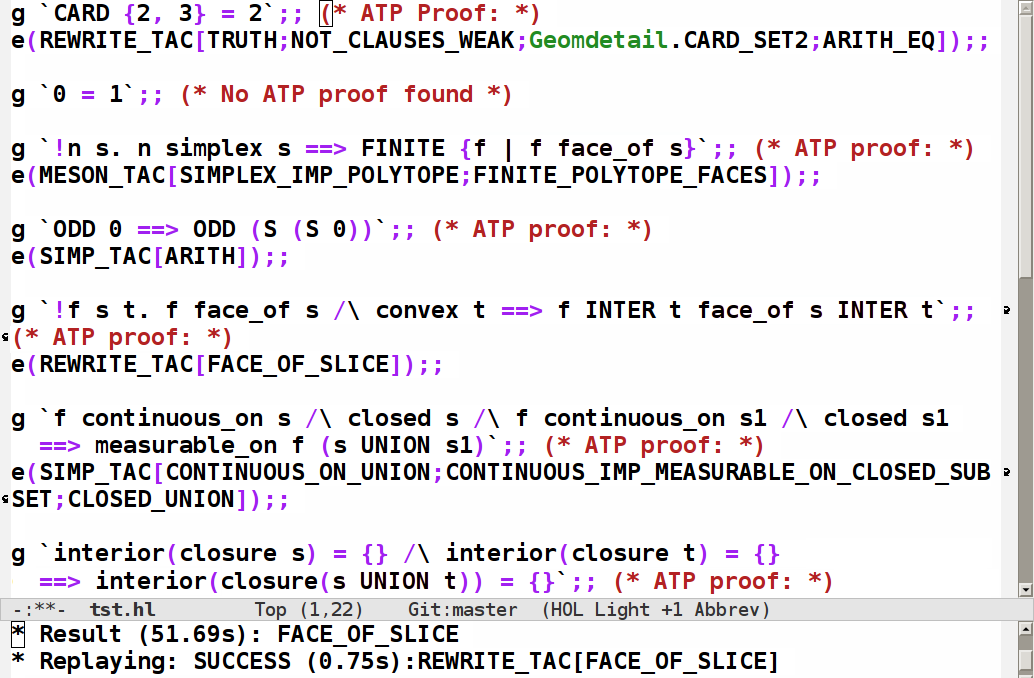}
\end{center}
\label{Emacs}
\end{figure}
Figure~\ref{Emacs} shows an Emacs session with several \HOLLight
goals.\footnote{A longer video of the interaction is at \url{http://mws.cs.ru.nl/~urban/ha1.mp4}} The online advisor has been asynchronously called on the 
goals, and just returned the answer for the fifth goal and
inserted the corresponding tactic call at an appropriate place in the
buffer. The relevant Emacs code (customized for the \HOLLight mode distributed with \Flyspeck) is available
online\footnote{\url{https://raw.github.com/JUrban/hol-advisor/master/hol-advice.el}}
and also distributed with the local \HH install. It is a
modification of the similar code used for communicating with the
\MizAR service from Emacs.

The simplest option (useful as a basis for  more sophisticated interfaces) is to interact with the service in command line, for example using \systemname{netcat}, as shown for two following two queries. The first query is solved easily by \texttt{INT_ARITH}, while the other requires nontrivial premise and proof search. 
\begin{small}
\begin{alltt}
\$ echo \textquotesingle{}max a b = &1 / &2 * ((a + b) + abs(a - b))\textquotesingle{}
  | nc colo12-c703.uibk.ac.at 8080
......
* Replaying: SUCCESS (0.25s): INT_ARITH_TAC
* Loadavg: 48.13 48.76 48.49 52/1151 46604

\$ echo \textquotesingle{}!A B (C:A->bool).((A DIFF B) INTER C=EMPTY) <=> ((A INTER C) SUBSET B)\textquotesingle{}
  | nc colo12-c703.uibk.ac.at 8080
 * Read OK
..............
 * Theorem! Time: 14.74s Prover: Z Hints: 32 Str: 
   allt_notrivsyms_m10u_all_atponly
 * Minimizing, current no: 9
.* Minimizing, current no: 6
 * Result: EMPTY_SUBSET IN_DIFF IN_INTER MEMBER_NOT_EMPTY SUBSET SUBSET_ANTISYM
\end{alltt}
\end{small}

\section{The Local Service Description}
\label{Local}

The service can be also
downloaded,\footnote{\url{http://cl-informatik.uibk.ac.at/users/cek/hh/}}
installed and used locally, for example when a user is working on a
private formalization that cannot be included in the public online
service.\footnote{The online service can already handle private developments that are not shown to the public.}
Installing the advisor locally proceeds analogously to the steps described in Algorithm~\ref{Setup1}.
Two passes are done through the user's
repository. In the first pass, the names of all the theorems available in the user's repository are exported,
together with their features (symbols, terms, types, etc., as
explained in Section~\ref{Features}). In the second pass, the
dependencies between the named theorems are computed, again
using the modified proof recording \HOLLight kernel
that records all the processing steps. Given
the exported features and dependencies, local advice system(s)
(premise selectors) are trained outside \HOLLight. Using the fast
sparse learning methods described in Section~\ref{Features}, this
again takes seconds, depending on the user hardware and the size of
the development.
The advisors are then run locally (as independent servers) to serve the
requests coming from \HOLLight. While the first pass is just a fast
additional function that can be run by the user at any time on top of
his loaded repository, the second pass now still requires full
additional processing of the repository. This could be improved in the
future by checkpointing the proof-recording image, as we do in the online server.

The user is provided with a tactic (\texttt{HH\_ADVICE\_TAC}) which
runs all the mechanisms described in the Section~\ref{Online} on the
current goal locally. This means that the functions relying on
external premise selection and ATPs are tried in parallel, together with a
number of decision procedures. The ATPs are expected to be installed
on the user's machine and (as in the online service) they are run on
the goal translated to the TPTP format, together with a limited number
of premises optimized separately for each prover. By default \Vampire,
\systemname{Eprover} and \Z are now run, using three-fold parallelization. 

The local installation in its simple configuration is now only trained
using the naive Bayes algorithm on the training data coming from the
\HOLLight proof dependencies and the features extracted with the
standard method. As shown
in~\cite{holyhammer}, the machine learning advice can be strengthened
using ATP dependencies, which can be also optionally plugged into the
local mode. Further strengthening can be done with
combinations of various methods. 
This is
easy to adjust; for example a user with a 24-CPU workstation can
re-use/optimize the parallel combinations from
Table~\ref{Combinations} used by the online service.

\subsection{Online versus Local Systems}
\label{Comparison}
The two related existing services are \MizAR and \Sledgehammer. \MizAR
has so far been an online service (accessible via Emacs or web
interface), while \Sledgehammer has so far required a local install
(even though it already calls some ATPs over a network). \HH started
as an online service, and the local version has been added recently to
answer the demand by some (power)users. The arguments for installing the service locally are mainly the option
to use the service offline (possibly using one's own large computing
resources), and to keep the development private. 
As usual, the local install will also require the tools involved to work
on all kinds of architectures, which is often an issue, particularly with
software that is mostly developed in academia.

As described in Section~\ref{Online}, the online service now runs 7
different AI/ATP instances and 4 decision procedures for each
query. When counting the individual ATP strategies (which may indeed
be very orthogonal in systems like \Vampire and \E), this translates
to about 70 different AI/ATP attempts for each query. If the demands grows,
we can already now distribute the load from the current 48-CPU server
to 112 CPUs by installing the service on another 64-CPU server.  The
old resolution-ATP wisdom is that systems rarely prove a result in
higher time limits, since the search space grows very fast. A more
recent wisdom (most prominently demonstrated by \Vampire) however is
that using (sufficiently orthogonal) strategy scheduling makes higher
time limits much more useful.\footnote{In~\cite{holyhammer}, the
  relative performance of \Vampire in 30 and 900 seconds is very
  different.} And even more recent wisdom is that learning in various
ways from related successes and failures further improves the systems'
chances when given more resources.
All this makes a good case for developing strong online computing
services that can in short bursts focus a lot of power on the user
queries, which are typically related to many previous problems. Also in
some sense, the currently used AI/ATP methods are only scratching the
surface. For example, further predictive power is obtained in \MaLARea~\cite{US+08}
by computing thousands of interesting finite models, and using
evaluation in them as additional semantic features of the
formulas. ATP prototypes like \MaLeCoP~\cite{UrbanVS11} can already
benefit from accumulated fine-grained learned AI guidance at every
inference step that they make. The service can try to make the best
(re-)use of all smaller lemmas that have been proved so
far (as in~\cite{Urban06-ijait}).  And as usual in machine learning, the more
data are centrally accumulated for such methods, the stronger the
methods become. Finally, it is hard to overlook the recent trend of
light-weight devices for which the hard computational tasks are
computed by large server farms (cloud computing).

\section{Conclusion and Future Work}
\label{Conclusion}

We believe that \HH is one of the strongest AI/ATP services currently
available.  It uses a toolchain of evolving large-theory methods that
have been continuously improved as more and more AI/ATP experiments and
computations have been recently done, in particular over the \Flyspeck
corpus. The combinations that jointly provide the greatest
theorem-proving coverage are employed to answer the queries with
parallelization of practically all of the components. The
parallelization factor is probably the highest of all existing ATP
services, helping to focus the power of many different AI/ATP methods
to answer the queries as quickly as possible. The content-encoding
mechanisms allow to re-use a lot of the expensive theorem-proving
knowledge computed over earlier projects and versions. And the
checkpointing allows reasonably fast update of existing projects.

At this moment, there seems to be no end to better premise selection,
better translation methods for ATPs (and SMTs, and more advanced
combined systems like \MetiTarski~\cite{AkbarpourP10}), better ATP
methods (and their AI-based guidance), and better reconstruction
methods.  Future work also includes broader updating mechanisms, for
example using git to not just add, but also delete files from an
existing project. A major issue is securing the server for more open
(perhaps eventually anonymous) uploads, and maybe also providing
encryption/obfuscation mechanisms that guarantee privacy of the
non-public developments.\footnote{The re-use performance obtained through content encoding suggests that just name obfuscation done by the client is not going to work as a privacy method.}
An interesting future direction is the use of the
service with its large knowledge base and growing reasoning power as a
semantic understanding (connecting) layer for experiments with tools
that attempt to extract logical meaning from informal mathematical
texts. Mathematics, with its explicit semantics, could in fact pioneer
the technology of very deep parsing of scientific natural language
writings, and their utilization in making stronger and stronger
automated reasoning tools about all kinds of scientific domains.

\bibliography{ate11}
\bibliographystyle{plain}

\end{document}